\pdfoutput=1

\documentclass[11pt]{article}

\usepackage{acl}

\usepackage{times}
\usepackage{latexsym, graphicx, amsmath, float, tabularx, subcaption}
\usepackage{amsfonts,amssymb}
\usepackage{xcolor}
\usepackage{array}
\usepackage{booktabs} 
\usepackage{threeparttable} 
\usepackage{cleveref}
\usepackage[T1]{fontenc}

\usepackage[utf8]{inputenc}

\usepackage{microtype}
\usepackage{hyperref}

\title{Evaluating Open-Domain Dialogues in Latent Space with Next Sentence Prediction and Mutual Information}

\newcommand*\samethanks[1][\value{footnote}]{\footnotemark[#1]}

\author{ 
    Kun Zhao\textsuperscript{1}\thanks{\quad \small Equal contribution.}\space,
    Bohao Yang\textsuperscript{2}\samethanks\space,
    Chenghua Lin\textsuperscript{2}\thanks{\quad \small Corresponding authors.}\;,
    Wenge Rong\textsuperscript{3,4},
    \\ \textbf{Aline Villavicencio}\textsuperscript{\textbf{2}}\textbf{,}
     \textbf{Xiaohui Cui}\textsuperscript{\textbf{1}} \samethanks\space\\ 
\textsuperscript{1} Key Laboratory of Aerospace Information Security and Trusted Computing, Ministry of\vspace{-0.5mm}\\
Education, School of Cyber Science and Engineering,  Wuhan University, China\vspace{-0.5mm} \\ 
\textsuperscript{2} Department of Computer Science, The University of Sheffield, United Kingdom\vspace{-0.5mm} \\
\textsuperscript{3} State Key Laboratory of Software Development Environment, Beihang University, China\vspace{-0.5mm} \\
\textsuperscript{4} School of Computer Science and Engineering, Beihang University, China\vspace{-0.5mm} \\
\texttt{
\{zhaokun, xcui\}@whu.edu.cn,  
}\vspace{-0.5mm} 
\texttt{
\ w.rong@buaa.edu.cn
}\vspace{-0.5mm} \\
\texttt{
\{byang27, c.lin, a.villavicencio\}@sheffield.ac.uk
}\vspace{-0.5mm} \\
}

\begin{document}
\maketitle

\begin{abstract}
The long-standing one-to-many issue of the open-domain dialogues poses significant challenges for automatic evaluation methods, i.e., there may be multiple suitable responses which differ in semantics for a given conversational context.
To tackle this challenge, we propose a novel learning-based automatic evaluation metric (\textbf{CMN}), which can robustly evaluate open-domain dialogues by augmenting Conditional Variational Autoencoders (\textbf{C}VAEs) with a Next Sentence Prediction (\textbf{N}SP) objective and employing Mutual Information (\textbf{M}I) to model the semantic similarity of text in the latent space. Experimental results on two open-domain dialogue datasets demonstrate the superiority of our method compared with a wide range of baselines, especially in handling responses which are distant to the golden reference responses in semantics.


\end{abstract}
\section{Introduction}

Open-domain dialogue generation is a prominent research direction in conversational AI due to a wide range of useful applications that it can facilitate, such as for personal digital assistants and customer service~\cite{sai_improving_2020,huang_challenges_2020,wang-etal-2021-fast,tang2023terminology}. While evaluating Natural Language Generation (NLG) systems is notoriously difficult, evaluation of open-domain dialogue generation introduces an extra layer of complexity, as a variety of responses can be generated, each semantically different and yet valid in the given context~\cite{li_diversity-promoting_2016, Gu2019DialogWAEMR, qiu_are_2019}.
For example, given the conversational context ``\textit{Iverson is my all-time favourite player.}'', responses such as ``\textit{He is my favourite player too.}'' or ``\textit{Yes, his quickness is amazing!}'' are both contextually relevant, yet semantically different.



Existing approaches for evaluating open-domain dialogue systems can be broadly divided into two different categories: reference-based and reference-free approaches. The reference-based metrics typically score a system by computing how similar an output response is compared to the \textit{gold-standard} reference. Popular metrics under this category may rely on surface-form similarity by counting the $n$-gram overlap between the response candidate and the reference (e.g., BLEU~\cite{papineni_bleu_2002}, ROUGE~\cite{lin_rouge_2004}, and METEOR~\cite{banerjee_meteor_2005}), or by calculating the similarity based on
embedding representations such as Embedding-Average~\cite{Wieting2016TowardsUP},
or even via high-dimensional representations learned for the response and the reference such as BERTScore~\cite{Zhang2020BERTScoreET}. One noticeable limitation of reference-based metrics is that they are reference centric and do not take the context of the conversation into consideration. Furthermore, due to the well-known one-to-many issue in open-domain dialogue~\cite{li_diversity-promoting_2016, Gu2019DialogWAEMR, qiu_are_2019}, a good response that matches well to its context could express significantly different semantics to its reference, for which the aforementioned metrics will be inadequate to handle.

To tackle the one-to-many issue, some works~\cite{Tao2018RUBERAU, Sinha2020LearningAU, Ghazarian2019BetterAE, Zhao2020DesigningPA} have proposed reference-free metrics to evaluate generated responses by measuring their similarity with the corresponding conversational context, by designing discriminative models trained on the context and the reference to judge whether the generated response matches the context well.
As these discriminative metrics are typically trained using a single relevant (aka. positive) response and multiple negative samples, \citet{sai_improving_2020} argue that such metrics should be trained with multiple relevant and irrelevant responses for any given context to allow for robust evaluation.
However, most existing datasets do not contain multiple references due to high cost of acquisition, rendering this recommendation impractical.

\citet{chan_enhancing_2021} take a different approach to the problem by evaluating generated responses in the latent space produced by Conditional Variational Autoencoders (CVAEs), as it can encode discrete text data into a smooth latent space~\cite{li-etal-2020-improving-variational,zhang2022improving}. Specifically, they proposed to use the prior distribution to approximate the conditional distribution for all the feasible responses to tackle the one-to-many issue with limited data. However,
there is no guarantee that the prior distribution can represent a rich set of feasible responses~\cite{Li2019ASE}.
\citet{zhang_mdd-eval_2022} proposed a self-training framework for multi-domain dialogue evaluation. The model performance was boosted by training on augmented datasets of four different domains, which are first automatically labelled by a teacher model and then followed by a manual annotation process.

To our knowledge, no prior works have attempted to model the intra-relation between a context and a response through the Next Sentence Prediction (\textbf{N}SP) task and Mutual Information (\textbf{M}I) directly, which can provide a strong signal for indicating the sequential and semantic dependencies between the context and response.

To tackle the one-to-many issue, we design
a reference-based automatic evaluation metric (\textbf{CMN}), which can robustly evaluate open-domain dialogues with a single gold-standard reference. Our method consists of a training stage and an evaluation stage. In the training stage, the \textbf{C}VAEs are augmented with a \textbf{N}SP objective~\cite{Devlin2019BERTPO}, which plays a crucial role in addressing the one-to-many issue in dialogue evaluation, especially when the semantics of the generated response are distant from the reference but still relate well to the context.

In the evaluation phase, we score a response candidate by calculating the \textbf{M}I of the context-response and response-reference pairs in the latent space, which are then combined through a weighting controlled by the NSP probability.
However, it is intractable to calculate MI directly as we only have access to samples instead of the prior and posterior distributions~\cite{paninski_estimation_2003, mcallester_formal_2018}.
To tackle this challenge, we propose to employ a contrastive learning method based on Noise Contrastive Estimation (NCE)~\cite{Gutmann2012NoiseContrastiveEO, logeswaran_content_2018} to calculate the lower bound of MI.
Overall, introducing the NSP objective and MI strengthens our model's ability to capture the sequential dependencies between the context and response, as well as to better leverage the information from references.

Experimental results on two open-domain dialogue datasets show the superiority of our method compared to a wide range of baseline metrics based on both Pearson and Spearman correlations with human annotations. In addition, we provide a detailed analysis of the effectiveness of our proposed method in solving the one-to-many issue in open-domain dialogue evaluation.
Our code is available at \textbf{\url{https://github.com/Bernard-Yang/CMN-ACL2023}}.

\section{Related Work}



\noindent\textbf{Reference-based metrics.}~~Reference-based metrics mainly compare the semantic similarity between a ground-truth reference and a generated response.
Representative metrics that calculate word overlap include BLEU~\cite{papineni_bleu_2002}, METEOR~\cite{banerjee_meteor_2005} and ROUGE~\cite{lin_rouge_2004}. 

Unlike metrics comparing the word overlap directly, embedding metrics first convert sentences into a high dimensional representation and calculate the semantic similarity between them. With the development of large-scale pre-training models, embedding metrics such as BERTScore~\cite{Zhang2020BERTScoreET} and BLEURT~\cite{Sellam2020BLEURTLR} have been shown to effectively enhance sentence representation. 
However, these automatic reference-based metrics cannot handle the well-known one-to-many problem in the open-domain dialogue.


\noindent\textbf{Reference-free metrics.}~~Existing reference-free metrics attempt to design discriminative models to solve the one-to-many issue by calculating the similarity between the context and the response candidate.
RUBER~\cite{Tao2018RUBERAU} is an unsupervised metric that calculates the similarity of the generated response with both the context and the response. MAUDE~\cite{Sinha2020LearningAU} employs a large-scale pre-trained model to convert sentences into hidden representations and leverage the temporal transitions between them.
\citet{sai_improving_2020} argued that such models should be trained on datasets containing multiple responses. However, most existing datasets only contain a single relevant reference and making this recommendation impractical.

EMS~\cite{chan_enhancing_2021} first attempted to utilise CVAEs to learn the reference information with limited data and approximate all feasible responses with the prior distribution. However, their model's prior distribution and sampled variables do not necessarily contain all the feasible response information for a given context, as EMS is only trained with a single reference. We propose a reference-based method by augmenting CVAEs with the NSP objective and employing MI to evaluate the response candidates.

\citet{zhang_mdd-eval_2022} tackled multi-domain evaluation by training a teacher model with human-annotated data in a particular domain. The model then labels the data from dialogue datasets in four other domains. This teacher-annotated data is then used to introduce a new evaluator, which can generalise across multiple domains.
However, this method requires human labelling and additional training data, which are not required by our method.

\section{Methodology}

\begin{figure*}[h]
  \centering
  \includegraphics[width=0.92\textwidth]{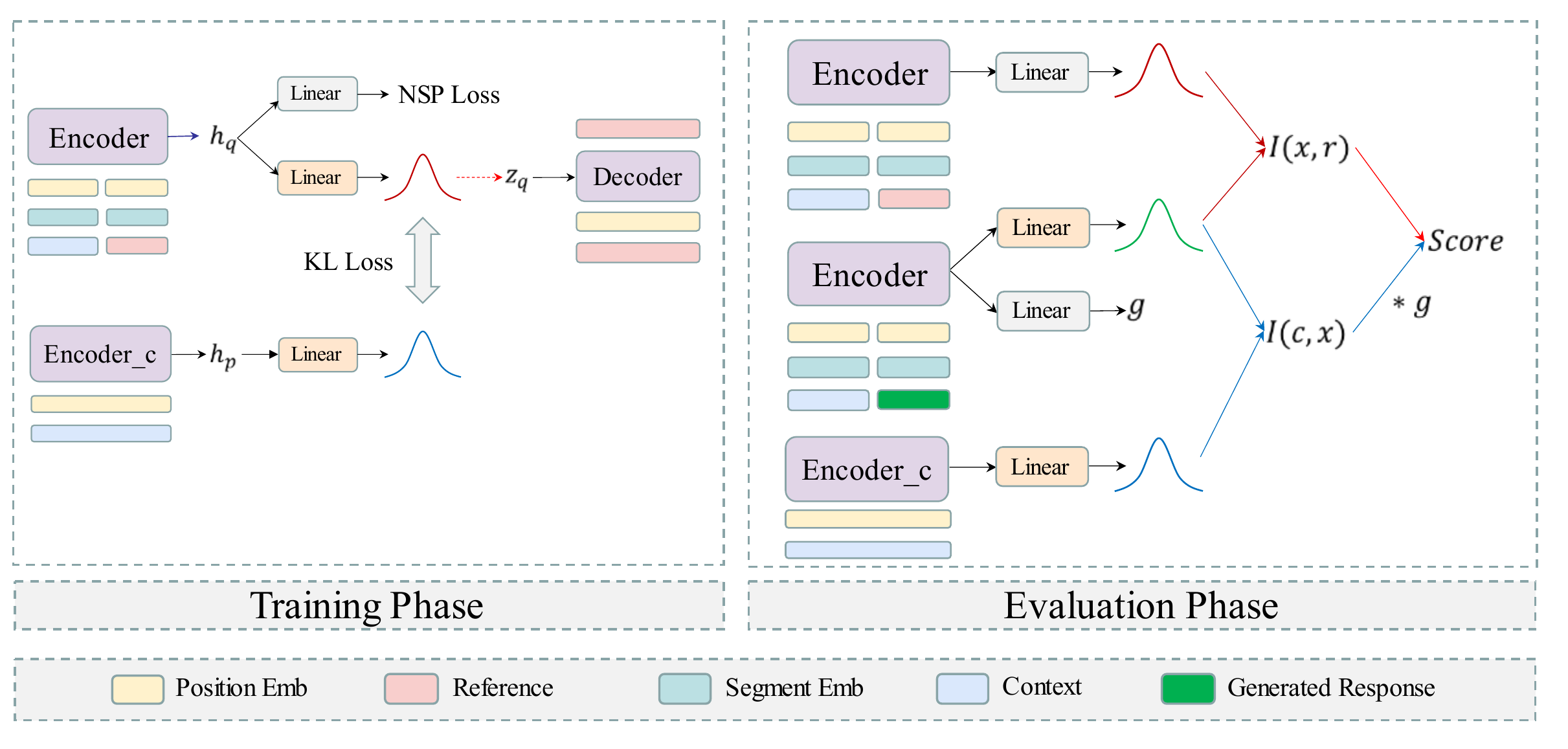}
  \caption{The architecture of the proposed model. The left part is the training phase, containing two encoders to represent the context-response pair and context. The NSP task is incorporated after the encoding of the context and response pair. The right part is the evaluation phase, in which the generated text is the response candidate and will be evaluated via the MI method during this stage. The Segment Embeddings are used for the NSP task.} 
  \label{fig: Model}
\end{figure*}





\subsection{Overall Architecture}

In this section, we describe the proposed automatic evaluation framework CMN in detail. As shown in Figure~\ref{fig: Model}, the overall architecture of CMN consists of two stages: training and evaluation.
The primary purpose of the training stage is to capture the dialogue information in the latent space, which is strengthened by incorporating the NSP objective into the CVAEs model.
In the evaluation stage, CMN evaluates the response candidates by calculating the MI of the context-response and response-reference pairs in the latent space, which are then combined through weighting with the NSP probability of the response candidate. 

\subsection{Training Stage}

The training process of our proposed method is illustrated in the left part of Figure \ref{fig: Model}. We employ two BERT encoders: the first is used to encode the context-reference pairs, and the second encodes the context only.
Formally, the encoding process is:
\begin{align}
       h_q = &\mathrm{Encoder}([c;r]) \nonumber \\ 
       h_p = &\mathrm{Encoder_c}(c)         \nonumber \\ 
       y = &\mathrm{Linear}(h_q)
\end{align}
where $h_q$ is the representation of the context-reference pair ($c$, $r$), and is used to learn the aggregated posterior distribution $q(z|c,r)$ of CMN. 
In contrast to EMS~\cite{chan_enhancing_2021} which does not model the order information of the context-reference pair, we introduce the segment embedding, which enables CMN to distinguish the order of the context and the reference. 
Finally, $y$ is the output of the NSP task, and $h_p$ is the representation of context, which is utilised to learn the prior distribution $p(z|c)$.

To address the one-to-many issue in open-domain dialogue evaluation, we introduce the NSP objective into the CVAEs' training process to enhance our model's discriminability of feasible responses given contexts. 
Introducing NSP leads to two different scenarios when training CMN. Specifically for the NSP task, we randomly replace the references fed to the encoder with the response from other conversations in the training set with a 0.5 probability, where the resulting context-response pairs are regarded as negative samples. Likewise, the contexts paired with the original references are positive samples. In terms of the input to the decoder, we use the original references (i.e. positive samples) during the whole training process, regardless of whether the inputs to the encoder are negative or positive samples.

\noindent\textbf{Training with positive samples.}~~When training with the positive samples, we add the NSP loss to the CVAEs' loss, where the NSP loss can be viewed as an additional regularisation, which enables the CVAEs model to capture the sequential dependencies between the context and response during the training stage. As a result, the posterior and prior distributions and the sampled latent variables will contain rich sentence order and pair matching information.
    \begin{align}
           &\mathcal{L}_{\text{train}}=\mathbb{E}_{q(z|c, r)}[\log p(r|c, z)]\nonumber\\&-\text{KL}(q(z|c,r)||p(z|c)) - \log p(y=1) 
    \end{align}
where $\mathbb{E}$ is expectation, $y=1$ indicates positive samples while $y=0$ indicates negative ones. The first term is the decoder reconstruction loss, the second term is the KL divergence, and the last term represents the cross entropy loss of the NSP task.

\noindent\textbf{Training with negative samples.}~~When training with the negative samples, we exclude the KL divergence loss of CVAEs, as it is undesirable to optimise the prior $p(z|c)$ to be close to the posterior $q(z'|c, r_{neg})$ of negative examples. 
 \begin{equation}
        \mathcal{L}_{\text{train}}=\mathbb{E}_{p(z|c)}[\log p(r|c, z)] - \log p(y=0)
 \end{equation}
Here $r$ is the reference from the datasets for guiding the decoder to generate reconstructed sentences. In addition, we use the prior distribution to sample $z$.



\subsection{Evaluation Stage}

In the evaluation stage, CMN learns to score a response candidate by calculating its MI with respect to the conversation context $c$ and the reference $r$ in the latent space. The representations of $c$ and $r$ are obtained in the training stage of CMN and contain rich sentence pair order and matching information. 

However, it is intractable to calculate MI directly as we only have access to the samples instead of the underlying distributions. 
To tackle this challenge, we employ InfoNCE, a contrastive learning method based on NCE
to  calculate the lower bound of MI between the latent variables of the two posterior probabilities $q(z|c, r)$ and $q(z|c, x)$ and prior probability $p(z|c)$ (see Figure~\ref{fig: Model} for illustration). Formally, the lower bound of MI is given as 
\begin{align}
\small
    & I(x,r) \geq {\mathbb{E}}_{(x, r)}[F(x, r)]+\log(N-1)\nonumber\\
    &-\mathbb{E}_x[\log \frac{1}{N-1} \sum_{r_{n} \in R_{neg}} e^{F(x,  r_{n})}] 
\label{InfoNCE}
\end{align}
where $x$ is the response candidate, $r$ is the ground-truth reference in the dataset, $r_{n}$ represents the negative response sampled from the negative set $R_{neg}$, which contains the references from other conversation turns, and $N$ is the number of negative samples. 

As the underlying posterior distributions are unknown, we first sample from each posterior probability to obtain latent variables $z_1$ and $z_2$, which contain the context-reference and the context-response sentence pairs information, respectively. 
The aforementioned sampling method, as well as the functions $F(x, r)$ and $F(x, r_n)$ in Eq.~\ref{InfoNCE}, are defined as follows:
\begin{align} 
    F&(x, r)=z_1 \cdot z_2 \nonumber\\ 
    &z_1 \sim q(z|c, r) \nonumber\\ 
    &z_2 \sim q(z|c, x)\nonumber\\
    F&(x, r_n)=z_1' \cdot z_2  \nonumber\\ 
    &z_1' \sim q(z|c, r_n) 
\label{Sample}
\end{align}
where $z_1$ and $z_2$ represent the positive latent variable samples while $z_1'$ represents the negative latent samples from the corresponding posterior distributions; $\cdot$ represents the dot product operation.
We can estimate the MI between response $x$ and reference $r$ (i.e. $I(x,r)$), as well as  the MI between context $c$ and response $x$ (i.e. $I(c,x)$), based on Eq.~\ref{InfoNCE} and Eq.~\ref{Sample}.

When calculating the final score for a candidate response, we also consider the NSP probability of the response candidate $x$ given conversational context $c$, in addition to the two MI values. 
The rationale is that InfoNCE might have difficulty measuring the semantic similarity between the response candidate $x$ and the reference $r$ when they are distant in semantics. The NSP probability acts as a natural weighting, informing the model of to what extent it should focus on $I(c,x)$, hence improving our method's robustness.
When feeding the context-response pair to the trained CVAEs in the evaluation stage, the NSP probability $g$ can be calculated according to the following formula:
\begin{equation}
    g=\sigma(\mathrm{Linear}(\mathrm{Encoder}([c;x])))
\end{equation}
where $\sigma$ is the activation function, and $g$ is the probability that response $x$ is predicted as the next sentence of context $c$. 
A higher value of $g$ means that the degree of dependence between context $c$ and response candidate $x$ is higher, and vice versa. 



Finally, we score a response candidate $x$ with Eq.~\ref{eq:final_score}. 

\begin{equation}
    \text{Score} = g * I(c,x) + I(x,r)
    \label{eq:final_score}
\end{equation}
The first term, $I(c,x)$, represents the semantic dependence of the context and the response candidate. In other words, it reflects how well the response candidate is related to the context. Thus using $g$ to multiply $I(c,x)$ controls the amount of information flowing from $I(c,x)$. In the second term, $I(x,r)$, we consider the semantic dependence of the response candidate and the reference based on their MI. 
Essentially, the relationship between $x$ and $c$, and that  between $x$ and $r$, can be considered simultaneously via Eq.~\ref{eq:final_score}, and the one-to-many problem can be handled directly.
\begin{figure*}[ht]
\small
\centering 
    \begin{subfigure}[]{0.35\textwidth}
	\includegraphics[scale=0.4]{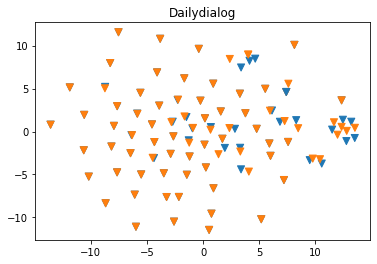}
	\caption{}
	\end{subfigure}
    \begin{subfigure}[]{0.4\textwidth}
	\includegraphics[scale=0.4]{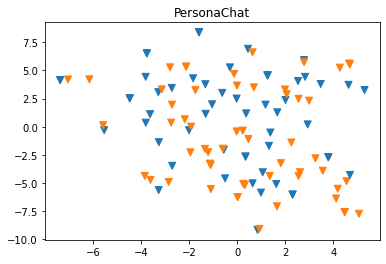}
	\caption{}
	\end{subfigure}%

	\begin{subfigure}[]{0.35\textwidth}
	\includegraphics[scale=0.4]{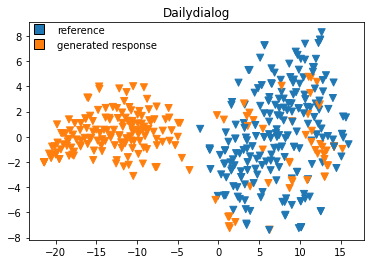}
    \caption{}
    \end{subfigure}%
    \begin{subfigure}[]{0.38\textwidth}
	\includegraphics[scale=0.4]{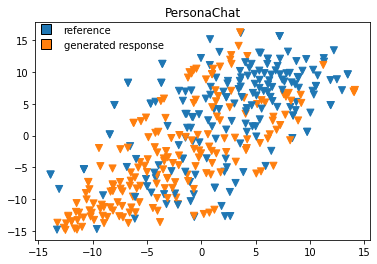}
	\caption{}
	\end{subfigure}%
\caption{T-SNE visualisation of the sentence representation of references and generated responses. (a) and (b) for the \textit{standard set}, and (c) and (d) for the \textit{diverse set}.}
\label{fig:label}
\end{figure*}
\section{Experimental Setup}
\subsection{Datasets}
To evaluate the effectiveness of our proposed automatic evaluation metric, we conduct experiments on two open dialogue datasets. 
The \textbf{PersonaChat} dataset~\cite{zhang_personalizing_2018} is a large persona-conditioned chit-chat style dialogue dataset which consists of 10,907 training dialogues, 1,000 validation dialogues, and 968 testing dialogues.  
The \textbf{DailyDialog} dataset~\cite{Li2017DailyDialogAM} is another widely-used large collection of human-human dialogues, consisting of a training set with 11,118 dialogues and validation and test sets with 1000 dialogues each.


\subsection{Baselines}
We choose the following two kinds of evaluation metrics as baseline methods:

\noindent\textbf{Reference-Based Metrics.}~~For the reference-based metrics, we use BLEU~\cite{papineni_bleu_2002}, ROUGE~\cite{lin_rouge_2004}, METEOR~\cite{banerjee_meteor_2005}, Embedding-Average~\cite{Wieting2016TowardsUP}, Vector-Extrema~\cite{Forgues2014BootstrappingDS}, Greedy-Matching~\cite{Rus2012ACO}, BERTScore~\cite{Zhang2020BERTScoreET}, and BLEURT~\cite{Sellam2020BLEURTLR}, which have been widely used in generative dialogue systems.

\noindent\textbf{Reference-free Metrics.}~~For the reference-free metrics, we compare with three learning-based methods, namely, RUBER~\cite{Tao2018RUBERAU}, MAUDE~\cite{Sinha2020LearningAU} and MDD-Eval~\cite{zhang_mdd-eval_2022}. Note that we were not able to compare with EMS~\cite{chan_enhancing_2021}, as their code is unavailable. It is also infeasible to re-implement their approach due to the lack of sufficient implementation details in the paper.

\subsection{Evaluation Set Construction} \label{Settings}

We follow the setting in Optimus~\cite{li_optimus_2020} to use BERT~\cite{Devlin2019BERTPO} and GPT-2~\cite{Radford2019LanguageMA} as the encoder and the decoder for our CMN framework, respectively. We set the dimension of the latent variable $z$ of CVAE to 32. 
In the evaluation phase, we follow~\citet{Zhao2020DesigningPA} to generate response candidates based on the testset of DailyDialog and PersonaChat using several widely applied dialogue generation systems, including Seq2Seq~\cite{sutskever_sequence_2014}, HRED~\cite{serban_hierarchical_2016}, and GPT-2~\cite{Radford2019LanguageMA}. 

After obtaining the generated response candidates, we construct an evaluation set consisting of a \textit{standard set}, in which the sample references and generated responses are similar in semantics (i.e., for the standard evaluation setting), and a \textit{diverse set}, in which the references and responses are distant in semantics (i.e., for the one-to-many setting). For the standard set, we collect 200 samples from both DailyDialog and PersonaChat that have the highest BLEU-1 scores between the reference and response among all the testing pairs. As our primary focus is to evaluate the model's performance under the one-to-many setting, we constructed a diverse set containing a larger number of samples (i.e., 600), by sampling from the testing pairs whose BLEU-1 scores are lower than 0.2. 
These sampled data have a balanced split between DailyDialog and PersonaChat.


\subsection{Human Annotation}


Evaluating the system performance requires measuring the correlation between the model prediction versus human evaluation scores. 
We recruited three human annotators to evaluate the evaluation set (i.e., the context-response pairs in our standard and diverse sets). 
All annotators hold at least a master's degree in Computer Science and have full professional proficiency in English. 

Specifically, annotators were asked to rate two aspects: \textbf{Appropriateness}, which measures the degree to which the output is appropriate within the given context, and \textbf{Coherence}, which assesses the extent to which the content of the output is presented in a well-structured, logical, and meaningful manner. These ratings were provided on a 1-5 Likert scale, with higher scores indicating better quality.
For each context-response pair, we then average the  Appropriateness and Coherence scores across all annotators  
to produce the final human annotation score. 
In the diverse set, 400 responses are rated as positive samples (4-5), while 200 are rated as negative samples (1-3). In contrast, all responses in the standard set are rated as positive samples since each response is semantically similar to the gold reference.

We examine the Inner-Annotator Agreement (IAA) using  inter-annotator Kappa ~\cite{doi:10.1177/001316446002000104}. 
The average IAA score between every pair of annotators for the Personachat dataset is 0.55, indicating a moderately strong level of agreement (0.4-0.6 range). On the other hand, the average IAA score for the DailyDialog dataset is 0.65, demonstrating a substantially strong level of agreement (0.6-0.8 range).
More details of the IAA scores can be found in Appendices \ref{sec:IAA}.

\begin{table*}[h]
\centering
\begin{threeparttable}[b]
\small
\begin{tabular}{lcc|cc}
             \toprule
& \multicolumn{2}{c|}{DailyDialog} & \multicolumn{2}{c}{PersonaChat} \\ \midrule
Metrics & Pearson's $\rho$ & Spearman's $\tau$ & Pearson's $\rho$ & Spearman's $\tau$ \\ \midrule

BLEU-1 & 0.0465 (0.6782) & 0.0049 (0.9652) & -0.0314 (0.8183) & -0.0372 (0.7856) 
        \\
BLEU-2 & 0.0497 (0.6577) & 0.0116 (0.9175) & -0.0601 (0.6597) & -0.0621 (0.6495)
        \\
BLEU-3 & 0.0462 (0.6803) & 0.0399 (0.7219) & -0.0431 (0.7525) & -0.0213 (0.8760)
        \\
BLEU-4 & 0.0796 (0.4770) & 0.0646 (0.5641) & -0.0149 (0.9134) & -0.064 (0.6395)
        \\ \midrule

ROUGE-1 & 0.0718 (0.5213) & 0.0304 (0.7861) & -0.0267 (0.8449) & 0.0678 (0.6193)
\\
ROUGE-2 & 0.0841 (0.4525) & 0.0645 (0.5651) & 0.0305 (0.8235) & 0.0291 (0.8315)
        \\
ROUGE-L & 0.0490 (0.6617) & 0.0285 (0.7992) & -0.0013 (0.9924) & 0.0834 (0.5413)
        \\ \midrule

METEOR & 0.0696 (0.5345) & 0.0946 (0.3977) & -0.102 (0.4543) & -0.1066 (0.4344)
        \\ \midrule
Embedding & & & &
        \\
Extrema & 0.1211 (0.2784) & -0.0021 (0.9853) & 0.0017 (0.9903) & 0.0814 (0.5510)
    \\
Greedy & 0.1117 (0.3176) & 0.0940 (0.4008) & 0.0949 (0.4866) & 0.0891 (0.5136)
    \\
Average & 0.1527 (0.1709) & 0.1199 (0.2835) & 0.1018 (0.4554) & 0.1124 (0.4096)
    \\ \midrule
BERTScore & 
0.0824 (0.4620) & -0.0076 (0.9457) & 0.1097 (0.4211) & 0.1724 (0.2038) \\
BLEURT & 0.1163 (0.2983) & 0.0940 (0.4008) & -0.1194 (0.3808) & -0.1143 (0.4016) \\
\midrule
RUBER & 0.0820 (0.4642) & 0.1560 (0.1616) & 0.0019 (0.9887) & -0.0329 (0.8095)
\\
MAUDE         & -0.1623 (0.1453) & -0.0145 (0.8974)
&  0.1353 (0.3201) & 0.1104 (0.4178)             \\
MDD-Eval & 0.1029 (0.3574) & -0.0667 (0.5516) & 0.1239 (0.3630) & 0.2502 (0.0629) \\
\midrule

Ours(w/o NSP) & 0.2292 (0.0383) & 0.2025 (0.0681) & 0.2585 (0.0544) & 0.3816 (0.0037) \\
Ours(w/o MI) & 0.0833 (0.4568) & -0.0537 (0.6316) & 0.1030 (0.4498) & 0.1530 (0.2601)\\

Ours  & \textbf{0.2446 (0.0268)} & \textbf{0.2211 (0.0459)} & \textbf{0.2656 (0.0479)} & \textbf{0.3971 (0.0024)}             \\
\bottomrule

\end{tabular}
\end{threeparttable}
\caption{Pearson and Spearman correlations with human judgements on the standard set. Figures in parentheses are p-values.}
\label{tab:similar_correlation}
\end{table*}

\section{Results}
In this section, we evaluate our model's performance on evaluating open-domain dialogues under both standard and diverse settings.

\begin{table*}
\centering
\begin{threeparttable}[b]
\small

\begin{tabular}{lcc|cc}
             \toprule
& \multicolumn{2}{c|}{DailyDialog} & \multicolumn{2}{c}{PersonaChat} \\ \midrule
Metrics & Pearson's $\rho$ & Spearman's $\tau$ & Pearson's $\rho$ & Spearman's $\tau$ \\ \midrule
BLEU-1 & 0.2953 (<0.0001) & 0.2635 (<0.0001) & -0.1533 (0.0361) & -0.1702 (0.0199)
\\
BLEU-2 & 0.2733 (<0.0001) & 0.2638 (<0.0001) & -0.1657 (0.0235) & -0.1810 (0.0132)
\\
BLEU-3 & 0.2496 (<0.0001) & 0.2691 (<0.0001) & -0.1654 (0.0237) & -0.1846 (0.0114)
\\
BLEU-4 & 0.2319 (<0.0001) & 0.2737 (<0.0001) & -0.1642 (0.0247) & -0.1790 (0.0142)
\\ \midrule
ROUGE-1 & 0.3275 (<0.0001) & 0.2865 (<0.0001) & -0.0057 (0.9382) & 0.0489 (0.5062)
\\
ROUGE-2 & 0.2698 (<0.0001) & 0.2761 (<0.0001) & -0.0340 (0.6441) & 0.0937 (0.2023)

\\
ROUGE-L & 0.3362 (<0.0001) & 0.2945 (<0.0001) & -0.0072 (0.9222) & 0.0476 (0.5178)
\\ \midrule
METEOR & 0.2948 (<0.0001) & 0.2858 (<0.0001) & -0.0293 (0.6908) & -0.0507 (0.4904)
\\ \midrule
Embedding & & & &
        \\
Extrema & -0.3589 (<0.0001) & -0.3524 (<0.0001) & -0.1010 (0.1690) & -0.0390 (0.5966)
    \\
Greedy & -0.1580 (0.0006) & -0.1408 (0.0023) & -0.0380 (0.6052) & 0.0113 (0.8776)
    \\
Average & -0.1350 (0.0034) & -0.1006 (0.0296) & -0.1093 (0.1364) & -0.0355 (0.6294)
    \\ \midrule 
BERTScore & 0.2591 (<0.0001) & 0.2251 (<0.0001) & 0.0345 (0.6391) & 0.0853 (0.2455)


\\
BLEURT & 0.2711 (<0.0001)) & 0.2063 (<0.0001)) & 0.1267 (0.0840) & 0.1858 (0.0109)\\
\midrule
RUBER & 0.1027 (0.0263) & 0.1714 (0.0002) & -0.0579 (0.4312) & -0.0592 (0.4206) \\
MAUDE & 0.0551 (0.2344) & 0.1782 (<0.0001) & 0.2640 (0.0003) & \textbf{0.3267 (<0.0001)} \\

MDD-Eval & 0.5567 (<0.0001) & 0.6160 (<0.0001) & 0.1264 (0.0848) & 0.2582 (0.0004) \\
\midrule
Ours(w/o NSP) & 0.5453 (<0.0001) & 0.5555 (<0.0001) & 0.2947 (0.0025) & 0.2224 (0.0022) \\
Ours(w/o MI) & 0.6183 (<0.0001) & 0.5946 (<0.0001) & 0.2769 (0.0001) & 0.1390 (0.0578) \\

Ours & \textbf{0.6325 (<0.0001)} & \textbf{0.6234 (<0.0001)} & \textbf{0.4000 (<0.0001)} & 0.2746 (0.0001) \\
\bottomrule

\end{tabular}
\end{threeparttable}
\caption{Pearson and Spearman correlations with human judgements on the diverse set.}

\label{tab:correlation}
\end{table*}

\subsection{Analysis of the Evaluation Set}

Before presenting the evaluation results, we first provide some validation analysis on our \textit{standard} and \textit{diverse} sets using embedding-based semantic similarity BERTScore. 
For the standard set, the averaged BERTScore is 4.7 for DailyDialog and 2.56 for Personachat. However, the scores are only 0.23 (DailyDialog) and 0.27 (Personachat) for the diverse set, indicating that the semantic similarity between the response candidates and the gold-standard references is low.

We further use T-SNE to visualise the sentence representations of the reference and generated response pairs.
As shown in Figure~\ref{fig:label} (a) and (b), the response candidates are similar to the references in the standard set, where the corresponding data points are either very close to each other or overlapping (e.g., there are seemingly more orange points in \ref{fig:label} (a) due to overlapping).
In contrast, the distributions of response candidates and references are more distinctive for the diverse set, as shown in Figure~\ref{fig:label} (c) and (d).  
In summary, the analysis shows that the standard and diverse sets are a good fit for our evaluation purposes.


\subsection{Model evaluation in the standard setting}



We compare our model with the baselines in terms of how well the evaluation scores generated by the model correlate with human judgments.

\textcolor{blue}{}
 


As shown in Table~\ref{tab:similar_correlation}, the $n$-gram baselines, including BLEU, ROUGE, and METEOR, achieve negative or weak positive correlations with human annotations on both datasets. The embedding-based approaches (including the ones using pre-trained models such as BERTScore) slightly outperform the $n$-gram baselines, except that BLEURT performs worse on the PersonaChat. 
In contrast, learning-based metrics give the strongest performance among all baselines. Specifically, MAUDE and MDD-Eval achieve similar performance on the PersonaChat, and both outperform RUBER. However, RUBER gives better performance than these two  metrics on DailyDialog.
Our model achieves the best overall performance in terms of both Pearson and Spearman correlations on both datasets.

We further conducted ablation studies to evaluate the effectiveness of the MI (w/o NSP) and the NSP (w/o MI) components by excluding the other component when inferring the final evaluation score.
It can be observed that CMN with the MI component alone (i.e., w/o NSP) gives better performance than the model variant with the NSP component only.
This suggests that MI is more effective than NSP in evaluating dialogues when the response candidates are similar to the references in semantics (i.e. the standard setting).

\subsection{Model evaluation in the one-to-many setting}
In another set of experiments, we evaluate our model performance in the one-to-many setting using the diverse set.

As shown in Table~\ref{tab:correlation}, Extrema, Greedy, and Average 
achieve a negative or weakly positive correlation with human annotation on both datasets. 
In contrast, the embedding-based metrics which use pre-trained models to represent sentences achieve much better results. For instance, both BERTScore and BLEURT achieve close to 0.25 for both Pearson and Spearman correlations on DailyDialog, although the performance is less strong on PersonaChat.

On the other hand, the word overlap metrics based on $n$-gram perform better than the above embedding-based metrics, with BLEU, ROUGE, and METEOR all having higher correlations than the embedding-based approaches. 
Nevertheless, the correlations of these metrics to human annotations are still relatively weak for both datasets.

For learning-based metrics, RUBER and MAUDE give weak positive correlations with human annotations on the DailyDialog dataset. However, RUBER gives a negative correlation with human scores on the PersonaChat.
MAUDE, on the other hand, performs the best on the PersonaChat dataset in terms of Spearman correlation (0.3267), which is higher than that of our method (0.2746). 
Overall, MDD-Eval gives the best performance among all baselines on DailyDialog, whereas MAUDE is the best baseline on PersonaChat. Nevertheless, our CMN model achieves the best overall performances on both datasets, giving the highest Pearson (0.6325) and Spearman (0.6234) correlations on DailyDialog and the highest Pearson (0.4000) correlations on PersonaChat.

Our ablation studies show that NSP is crucial in evaluating dialogues when there is a significant difference between references and responses in semantics (i.e., the diverse setting). By introducing NSP, our model can effectively capture the contextual dependencies between the conversational context and the generated responses, and thus can better handle the one-to-many issue in open-domain dialogue evaluation.



\begin{table}[h]
\small
\begin{tabular}{cccl}
\toprule
\multicolumn{1}{l}{\textbf{Context:}} & \multicolumn{3}{l}{\begin{tabular}{p{5cm}}What do you need?\end{tabular}} \\ \midrule
\multicolumn{1}{l}{\textbf{Reference:}} & \multicolumn{3}{l}{\begin{tabular}{p{5cm}}I need to use the internet .\end{tabular}} \\ \midrule
\multicolumn{1}{l}{\textbf{Response:}} & \multicolumn{3}{l}{\begin{tabular}{p{5cm}}I think I need a deck that plays well with this.\end{tabular}} \\ \midrule
Human & BLEU & MAUDE & \multicolumn{1}{c}{RUBER} \\ 
4.66 & 0.90 & 4.81 & \multicolumn{1}{c}{0.85} \\ \hline
BERTScore & BLEURT & MDD-Eval & \multicolumn{1}{c}{Ours} \\ 
1.90 & 1.34 & 0.53 & \multicolumn{1}{c}{4.47} \\ 

\midrule
\multicolumn{1}{l}{\textbf{Context:}}   & \multicolumn{3}{l}{\begin{tabular}{p{5cm}}Do you like the outdoors?\end{tabular}}                                                                                                                            \\ \hline
\multicolumn{1}{l}{\textbf{Reference:}} & \multicolumn{3}{l}{\begin{tabular}{p{5cm}}I like taking my dogs hiking. What do you like to do for fun?\end{tabular}} \\ \hline
\multicolumn{1}{l}{\textbf{Response:}} & \multicolumn{3}{l}{\begin{tabular}{p{5cm}}I do. I love to hike.\end{tabular}}                                                                                                                     \\ \midrule
Human                          & BLEU                                               & MAUDE                                             & \multicolumn{1}{c}{RUBER}                                          \\ 
5.0         & 1.25      & 4.94                                               & \multicolumn{1}{c}{0.76}                                            \\ \hline
BERTScore & BLEURT  & MDD-Eval  & \multicolumn{1}{c}{Ours}  \\
1.66      & 2.12    & 3.93   & \multicolumn{1}{c}{4.26} \\
\bottomrule
\end{tabular}
\caption{Samples from DailyDialog and PersonaChat dataset.}
\label{tab:case study1}
\end{table}

\subsection{Case Studies}
For qualitative analysis, we show two cases of our experiment in Table~\ref{tab:case study1}. Each case shows the conversational context as well as the corresponding gold-standard reference and the generated response. We compare our evaluation score with five different baselines. To simplify the comparison, we normalise all scores to a range of 1-5 to be consistent with the Likert scale of human evaluation. 
Note that the normalisation is applied to the case study only, rather than performed in our main  experiments.
In the first case, the generated response is relatively similar to the reference, whereas the reference and response are very different in the second case.
For both cases, our CMN gives very similar scores to the human scores.
More examples are provided in Appendices~\ref{sec:app_case}.



\section{Conclusions}

In this paper, we propose a novel learning-based automatic evaluation metric which can robustly evaluate open-domain dialogue by augmenting CVAEs with an NSP objective and employing MI to model the semantic similarity of text in the latent space. 
Experimental results on two open-domain dialogue datasets show that our CMN model outperforms a wide range of baseline methods in terms of both Pearson and Spearman correlations 
with human annotation scores, and is superior in dealing with the one-to-many issue in open-domain dialogue evaluation.

\section*{Ethics Statement}

In this paper, we propose a new automatic evaluation metric CMN to evaluate the open-domain dialogue system. The positive impact of CMN is that it can deal with the one-to-many problem in the open-domain dialogue evaluation metrics. The negative impact lies in that the CMN may potentially give a high score to potentially inappropriate or offensive responses in some extreme cases. Consequrntly, the content of such training datasets should be assessed before training the CMN.
\section*{Limitations}

Although our proposed method performs well in evaluating the open-domain dialogue systems, it also has some limitations. Our method identifies the dependencies between context and response. However, according to \citet{Howcroft2020TwentyYO}, human-evaluated metrics can contain a variety of attributes whilst we only identify the large-scale dependencies of semantics and do not disentangle the texts into the attributes of human-evaluated metrics. In the future, we will conduct disentanglement studies to disentangle the text into various attributes to optimise our model and further improve the interpretability of text evaluation methods based on these disentangled attributes.

\bibliography{anthology,custom, zotero_references}
\bibliographystyle{acl_natbib}

\appendix
\label{sec:appendix}
\section{Appendices}
\subsection{Case Studies}
\label{sec:app_case}
We demonstrate more examples in Table~\ref{tab:case study2}, which shows the response and the reference conditioned on the same conversational context from the PersonaChat dataset. We compare our matching score with five different baselines.
We notice that the matching score of our method correlated well with the human annotation score compared with other baselines.

\begin{table}[h]
\small
\begin{tabular}{cccl}
\toprule
\multicolumn{1}{l}{\textbf{Context:}}   & \multicolumn{3}{l}{\begin{tabular}{p{5cm}}I love nature ! i'm going camping tomorrow night\end{tabular}}                                      \\ \hline
\multicolumn{1}{l}{\textbf{Reference:}} & \multicolumn{3}{l}{\begin{tabular}{p{5cm}}It is too cold here to go camping.\end{tabular}}                                                    \\ \hline
\multicolumn{1}{l}{\textbf{Response:}} & \multicolumn{3}{l}{\begin{tabular}{p{5cm}}That sounds fun . i like to go to the beach.\end{tabular}}                        \\ \midrule
Human  & BLEU  & MAUDE & \multicolumn{1}{c}{RUBER} \\ 
4.5    & 0.8   & 4.96  & \multicolumn{1}{c}{0.44}  \\ \hline
BERTScore & BLEURT  & MDD-Eval  & \multicolumn{1}{c}{Ours}  \\
2.01                           & 1.71                                 & 2.90  &  \multicolumn{1}{c}{4.21}  \\ \midrule
\multicolumn{1}{l}{\textbf{Context:}}   & \multicolumn{3}{l}{\begin{tabular}{p{5cm}}I have a cat. His name is spook. What about you?\end{tabular}}                                      \\ \hline
\multicolumn{1}{l}{\textbf{Reference:}} & \multicolumn{3}{l}{\begin{tabular}{p{5cm}}I have a turtle. I named him leo.\end{tabular}}                                                                                                             \\ \hline
\multicolumn{1}{l}{\textbf{Response:}} & \multicolumn{3}{l}{\begin{tabular}{p{5cm}}I've a dog, but he has black and white eyes, what about you?\end{tabular}}                        \\ \midrule
Human                          & BLEU                                               & MAUDE                                             & \multicolumn{1}{c}{RUBER}                                          \\ 
4.5        & 0.35      & 4.98                                           & \multicolumn{1}{c}{0.61}                                            \\ \hline
BERTScore & BLEURT  & MDD-Eval  & \multicolumn{1}{c}{Ours}  \\

1.29     & 1.89   &0.53    & \multicolumn{1}{c}{4.16}    \\ 

\bottomrule
\end{tabular}
\caption{Three samples from DailiDialog and PersonaChat dataset.}
\label{tab:case study2}
\end{table}

\subsection{Inter-Annotator Agreement (IAA)}
\label{sec:IAA}

We use cohen's kappa~\cite{doi:10.1177/001316446002000104} to examine the IAA between every two annotators and demonstrate our IAA in Table \ref{tab:IAA}. All the IAA scores of the Personachat dataset are higher than 0.4, which indicates that the annotators reached a moderately strong level agreement (0.4-0.6) or a substantially strong level agreement (0.6-0.8). Besides, the IAA scores of the DailyDialog dataset can reach a substantially strong level. The above IAA results indicate that the annotated data by different annotators are reliable.

\begin{table}[h]
\small
\centering
\begin{tabular}{cccc} 
\toprule
 \multicolumn{1}{l}{\textbf{Annotator}} & \multicolumn{3}{c}{{Cohen's Kappa}}   \\ \hline   
\multicolumn{4}{c}{DailyDialog}  \tabularnewline
\midrule
  \multicolumn{1}{l}{} & Annotator1 & Annotator2 & Annotator3 \\ \hline   
 \multicolumn{1}{l}{Annotator1}       &- &0.6896 & 0.6035\\ \hline   
 \multicolumn{1}{l}{Annotator2}      &0.6896 & - & 0.6434 \\ \hline   
 \multicolumn{1}{l}{Annotator3}      & 0.6035 & 0.6434 & - \\  
\toprule
  \multicolumn{4}{c}{PersonaChat} \tabularnewline
\midrule
  \multicolumn{1}{l}{\textbf{Annotator}} & Annotator1 & Annotator2 & Annotator3 \\ \hline   
 \multicolumn{1}{l}{Annotator1}      & - &0.4496 & 0.5547  \\ \hline  
 \multicolumn{1}{l}{Annotator2}    & 0.4496 &- &0.6315  \\ \hline   
 \multicolumn{1}{l}{Annotator3}      & 0.5547 & 0.6315 & -   \\  
\bottomrule
\end{tabular}
\caption{Inter-Annotator agreement (IAA)}
\label{tab:IAA}
\end{table}

\end{document}